# Fine-tuning Large Language Models with Generative Reward Modelling

**Zhang Ze Yu**
School of Computing
National University of Singapore
`zhan1130@comp.nus.edu.sg`

**Zhang Hui**
School of Computing
National University of Singapore

**Lau Jia Jaw**
School of Computing
National University of Singapore

**Bryan Kian Hsiang Low**
School of Computing
National University of Singapore

## Abstract

Reinforcement Learning with Human Feedback (RLHF) has been demonstrated to significantly enhance the performance of large language models (LLMs) by aligning their outputs with desired human values through instruction tuning. However, RLHF is constrained by the expertise and productivity limitations of human evaluators. A response to this downside is to fall back to supervised fine-tuning (SFT) with additional carefully selected expert demonstrations. However, while this method has been proven to be effective, it invariably also leads to increased human-in-the-loop overhead. In this study, we propose another alternative approach: Reinforcement Learning with Generative Adversarial Feedback (RLGAF) to RLHF and SFT, which uses a generative adversarial training style to enable the LLMs to learn useful human expert demonstrations without being directly exposed to the training examples, thus enabling good generalization capabilities while preserving sample efficiency. Our preliminary findings indicate that RLGAF can help align LLMs outputs with competitive performance against RLHF and SFT, while not suffering from their respective inherent restrictions, suggesting promising avenues for further research on automating AI alignment.

## 1 Introduction

Large Language Models (LLMs) hold immense potential for enhancing productivity across a wide range of use cases. However, it is crucial to acknowledge the risks associated with producing undesirable outputs, such as inaccurately generated facts and harmful suggestions Beutel et al. [2023], Feldman et al. [2023], Manakul et al. [2023], Bender et al. [2021], Bommasani et al. [2021]. OpenAI has adopted reinforcement learning with human feedback (RLHF) as the primary method for aligning the outputs of LLMs Ouyang et al. [2022]. This approach integrates human preferences into LLMs by treating the model as a reinforcement learning agent to maximize its reward when the output sequences are aligned with human preferences, thereby mitigating misalignment issues in their outputs Liu [2023], Christiano et al. [2017], Lin et al. [2020], Stiennon et al. [2020]. Despite its visible improvements, RLHF necessitates human involvement to provide feedback on LLMs' outputs, which can be both time and resource-intensive. Due to this reason, RLHF potentially suffers from scalability issues. With the increasing diversity of data, human annotators are required to handle a growing workload, which may be proved too much to keep up with. Moreover, the effectiveness of RLHF is constrained by the expertise of human annotators. Certain labeling tasks have proven to be too complex or necessitate niche expertise, posing challenges for many annotators to score accurately, given the intricate nature of some tasks, the ambiguity in certain subject areas or the plausible but deceptive outputs the model can give Aiyappa et al. [2023]. Hence, there is a need to explore other methods to align the LLMs.

As a response to the aforementioned downsides, Zhou et al. [2023] proposed to replace RLHF with supervised fine-tuning (SFT) during the instruction tuning stage, where a set of carefully curated prompts and expert demonstration responses are used to further fine-tune the LLMs to output the intended responses. This approach bypasses the need to do reward modeling as well as reinforcement learning, arguably simplifies the alignment process. However, since LLMs are trained specifically to imitate the given demonstration, exposing it directly to the expert demonstration could potentially lead to superficially mimicking the form of the response only. As an example, when the expert response is hedged or uninformative (e.g., 'not sure'), the model might only mimic the form (hedging or just output 'I don't know') even if it does know the answer. On the other hand, when the expert response is some definitive answer to the prompt, the model might be led to output a definitive answer even if it has no knowledge pertaining to the specific topic. As such, the poor handling of uncertainty estimation of SFT can directly lead to exacerbating hallucinations. Furthermore, the human effort needed for curating a high-quality expert demonstration dataset that meets the variety and form requirements is still non-negligible, even if the expert data size is comparatively much smaller than that needed in RLHF.

In this work, we propose using Reinforcement Learning with Generative Adversarial Feedback (RLGAF) to jointly perform reward modelling and LLMs instruction tuning without separately training a reward model or exposing the model directly to the fine-tuning data. Generative adversarial training is employed to train an LLM and utilizes a discriminator to assess the quality of the outputs generated by the generator—in this case, the LLM we aim to align. The discriminator itself is another LLM, tasked with providing a score for the generated sequence from the first LLM (i.e., the generator). In essence, the discriminator takes the first LLM's output as input and returns a scalar value as output. In contrast to RLHF's reward model (RM), the discriminator does not require supervised learning with human-labeled (i.e., ranked) data. Instead, it undergoes training together with the generator in an alternating manner. The training process mirrors that of a typical GAN, with the discriminator receiving positive examples (i.e., expert demonstrations) and negative samples from the generator's outputs. This approach allows the discriminator to learn reward modeling as the generator produces increasingly better outputs. On one hand, similar to GANs, this method facilitates improved learning signals for the generator, even in the absence of explicit ranking for output quality at different levels as a pre-trained RM does Ouyang et al. [2022]. On the other hand, unlike SFT, since the generator is not directly exposed to the expert data during the instruction tuning stage, it lowers the risk of overfitting which could potentially lead to worse hallucination.

## 2 Preliminaries and Proposed Approach

The core idea of RLGAF is to approach reward modelling with a GAN architecture, where the discriminator takes the place of both human and the reward model, and provides feedback to the generator. This creates a pseudo-online reward modelling setting, where half of the training data for the discriminator are fresh (i.e., generated by the generator) at each round of generator training. This mitigates various issues from RLHF's completely off-line reward modelling approach such as misgeneralization and off-distribution Levine et al. [2020], Gao et al. [2023], while not at the cost of drastically increasing the human-in-the-loop effort incurred in a completely online reward modelling setting where human annotators need to rank the output of the LLM after every round of its training Casper et al. [2023].

Generative adversarial networks (GANs) have been used extensively in image generation fields Pan et al. [2019], Aggarwal et al. [2021], Gui et al. [2021]. This architecture generally works well for image-related tasks, where the inputs and outputs are of continuous values. However, text generation involves the use of discrete tokens. In particular, the generated response passed to the discriminator are in the form of discrete tokens. This prevents the computed loss to be back-propagated from the discriminator to the generator. Furthermore, the discriminator can only provide a score after the whole sequence has been generated since the intermediate score does not represent the quality of the whole sequence Yu et al. [2017].

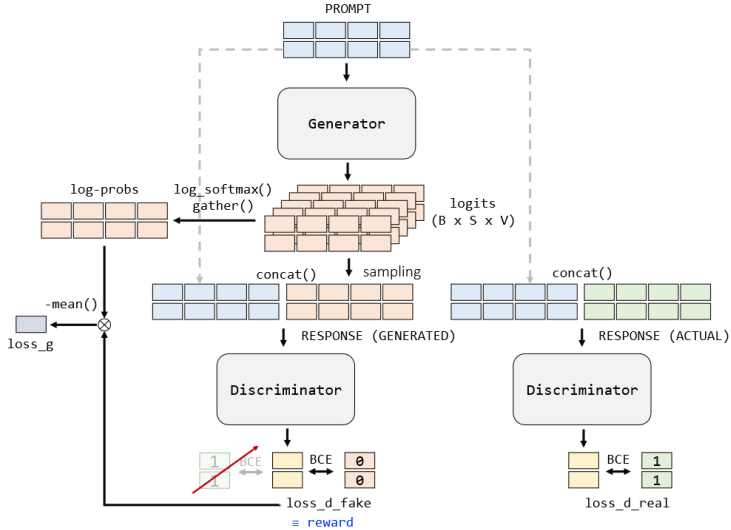

Figure 1: GAN model architecture for policy gradient method. 'loss_g' denotes the generator loss, 'loss_d_fake' denotes the discriminator loss on label '0' data, 'loss_d_real' denotes the discriminator loss on label '1' data.

In this work, we mainly explored two policy gradient methods on full-length generated texts for gradient updates for the aforementioned challenges in using GANs on text generation tasks.

### 2.1 Policy Gradient

Text generation can be seen as a sequential decision-making process where each word depends on previously generated words. Naturally, we can formulate the text generation task in a way where the state is the words generated so far and the action is the choice of the next word; the objective is to maximise the reward based on the task of interest. Therefore, we can use reinforcement learning to train the generator in a GAN.

One related prior work is SeqGAN: Sequence Generative Adversarial Nets with Policy Gradient Yu et al. [2017]. SeqGAN solves the problem of unable to back-propagate discriminator gradient to the generator by using policy gradient method for training. Meanwhile, it employs Monte Carlo Tree Search (MCTS) to roll out samples from all possible complete sentences as actions taken, which solves the problem that incomplete sentences do not have well-defined rewards. However, the focus of this work is text generation. While SeqGAN appears to mimic the style of the target text, the author did not investigate if it is equally effective for alignment purposes.

The generator and discriminator are trained in an interleaving manner. Real data and generated data are passed to the discriminator for training, with real data labeled as '1' and generated data as '0'. In generator training, full-length sequences are first generated by the generator, then we iterate through each word position and sample N new sentences with the roll-out model (generator) from all the already generated words; the full-lengths sentences sampled are then passed to the discriminator to calculate the reward. The average reward of all sampled sequences is calculated to update the generator. For implementation, SeqGAN uses LSTM for the generator and CNN for the discriminator.

Despite Yu et al. [2017] has shown GANs training can enable language models to learn to generate outputs of a similar form as the data from a target distribution, this does not guarantee the generated data do not just superficially mimick the form of the training data. In this work, our focus is on evaluating if GANs training can indeed fulfill the criteria required for alignment. We attempt to use LLMs as the base model and implement RLGAF with them in two different ways:

### 2.1.1 Monte Carlo Policy Gradient (REINFORCE)

This is a classic policy gradient method Sutton and Barto [2018]. From the policy gradient theorem

$$\nabla_\theta J(\theta) = \mathbb{E}_{\tau \sim \pi_\theta} \left[ \sum_{t=0}^{T-1} \nabla_\theta \log \pi_\theta(a_t|s_t) \sum_{t'=t}^{T-1} r(s_{t'}, a_{t'}) \right], \quad (1)$$

we know that we can estimate the gradient of the RL agent policy with Monte Carlo method as follows Sutton and Barto [2018]:

$$\hat{g}(\theta) = \frac{1}{N} \sum_{i=1}^{N} \sum_{t=0}^{T-1} \nabla_\theta \log \pi_\theta(a_t^{(i)}|s_t^{(i)}) \sum_{t'=t}^{T-1} r(s_{t'}^{(i)}, a_{t'}^{(i)}). \quad (2)$$

In our first implementation of RLGAF, we choose to use this method to estimate policy gradient and iteratively optimize the RL agent's policy through gradient ascent to train the generator.

### 2.1.2 Proximal Policy Gradient (PPO)

In the second version of RLGAF implementation, we substitute the reinforcement learning component (i.e., generator training) with the same approach in the InstructGPT paper Ouyang et al. [2022]:

$$\text{objective}(\phi) = \mathbb{E}_{(x,y) \sim D_{\pi_\phi^{\text{RL}}}} \left[ r_\theta(x,y) - \beta \log \frac{\pi_\phi^{\text{RL}}(y|x)}{\pi^{\text{SFT}}(y|x)} \right] + \gamma \mathbb{E}_{x \sim D_{\text{pretrain}}} \left[ \log(\pi_\phi^{\text{RL}}(x)) \right]. \quad (3)$$

This formulation essentially adopts the original PPO objective Schulman et al. [2017] specifically into the context of aligning LLMs. In particular, it treats the pretrained LLM of interest as an RL agent, and one entire output sequence as an action. Its policy over action space chooses some sequence out of all the possible complete sequences the LLM might be outputting. The objective of alignment is to choose the more desirable sequence out of all possible sequences.

## 3 Form and Sentiment Alignment Experiments

### 3.1 Methods

In our experiments, we use pre-trained LLMs for our generator and discriminator. The pre-trained models provide a baseline language modeling and generative capabilities, and the goal here is to fine-tune these pre-trained models to generate appropriate responses to input prompts. We use the same model architecture for both the generator and discriminator in order to match the generated tokens with the discriminator's token vocabulary. The training workflow is described in Table 1 and visualised in Figure 1.

Just as in the case of RLHF, RLGAF primarily aimes at aligning the model with a specific domain it is being trained on, rather than significantly improving its output quality. As we use GPT-2 (124 million parameters) as our base model in this experiment, we do not anticipate our aligned model to perform better than the base model. Rather, our goal is to train a model that performs better in the specific domain we are targeting by using RLGAF without requiring human intervention as in RLHF. In our case, we use RLGAF to help us pick the most desirable outputs GPT-2 base model can generate for Question-Answering domain and Movie Review Sentiment Analysis domain.

For Question-Answering domain, we train our model on the SQUAD 2.0 dataset Rajpurkar et al. [2016]. This dataset contains contexts, questions and answers about a wide range of topics, and our objective is to train our RLGAF model to generate answers with sentence structures that are similar to the ones in the training dataset. However, we realize that the limitations of GPT-2 may prevent our RLGAF model from generating answers that are fully accurate or semantically meaningful.

| **Training the discriminator (real data)** |
| --- |
| 1. The prompt and actual response are passed as inputs to the discriminator. |
| 2. The discriminator outputs a score which, after applying a sigmoid function, represents the probability that the input came from the real dataset. |
| 3. A binary cross-entropy (BCE) loss is computed between the score and expected label '1' for the real dataset. |
| **Training the discriminator (generated data)** |
| 1. The prompt is passed as inputs to the generator, which generates a response. |
| 2. Logits are detached to ensure the gradients do not back-propagate to the generator. |
| 3. The response tokens, along with the prompt, are passed as inputs to the discriminator. |
| 4. The discriminator outputs a score for the generated response. |
| 5. BCE loss is computed between the score and expected label for generated data labeled '0'. |
| 6. The discriminator's overall loss is the average between the loss for the real data and the loss for the generated data. |
| 7. The loss is back-propagated to update the discriminator's weights. |
| **Training the generator** |
| 1. Similar to training the discriminator, a forward pass is performed on the generator and the discriminator. |
| 2. However, the expected label for computing BCE loss is now '1', since the generator's goal is to fool the discriminator and seeks to maximize the discriminator's loss. |
| 3. The loss is back-propagated to the generator's weights via the discriminator. |

Table 1: Procedural steps for RLGAF training.

For Movie Review Sentiment Analysis domain, we train our model on the IMDB dataset. This dataset contains movie reviews labeled with positive or negative sentiment. We aim to train our LLM to generate movie reviews with only positive sentiments. We run RLGAF training so that the discriminator rewards the reviews that contain positive sentiment, and punish the reviews that contain negative sentiment.

### 3.2 Evaluation and Discussion

#### 3.2.1 Monte Carlo Policy Gradient

With reference to Figure 4, we observe consistent improvement in both the discriminator and generator during the training iterations. Nevertheless, improvement slows down considerably after iteration 200 and stagnant thereafter. After the training, we find that our model produces sentences with shorter lengths more frequently compared to the GPT-2 base model, similar to the short answer found in the SQUAD 2.0 dataset but in a more complete sentence manner. However, the quality of the generated sentences in terms of semantics meaning is not as good as that of the GPT-2 base model.

Table 8 shows some samples generated from both our Policy Gradient Method model and GPT-2 base model using the same hyperparameters. While the base model generated longer and more complete sentences, as expected from a sentence completion model, our model generated shorter sentences in multiple lines, sometimes with repetitive responses. This suggests that while the generator was not able to output the correct keywords from the context given, it has learned that the correct outputs are supposed to be in just a few words or a short phrase as opposed to lengthy complete sentences. This is expected, as GPT-2 based question answer model requires very different methods to train (outputting correct index of the starting and ending keywords as opposed to text generation). Nevertheless, it is evident that RLGAF training approach can successfully inform part of the feature that the desired outputs are supposed to have.

Table 2: Generating outputs with only positive sentiment

| Prompt | Generated Response | Sentiment |
| --- | --- | --- |
| May I please | , as I write this,I would like to warn you before spreading this cliché of film to our nation,««< 25 | negative |
| I'm sure that | with the way the film has been rated a B in its day, it would have been better. This film is definitely deserved««< | positive |
| I ve finished | watching this film, I highly, highly recommend you to anyone who is a fanter of movie, this movie«««< | positive |
| Holy freaking God | !)<br /><br />A movie which focuses on its protagonist (and, unfortunately, his directorial debut), is perhaps only a worthy movie.«« | positive |
| Quote: theurgist | : Anyone remember when there was one film made which was so successful?<br /««««« 50 MP | positive |
| Well...now | that I'm done, I must say that this movie is an excellent example of one-acters movie, that it is so< "< MP | positive |
| **SPOILERS** Redicules slasher | are a thing. That said, in this film. when one movie.«< 25 WORST Were | unclear |
| Not that I | know what some people have been saying. This movie scared the hell out of me. This is really one movie that«« | positive |
| Due to budget | , and director Michael Lombardo's style of directing, this film, is probably the best-titled film of all time.It< I« | positive |
| I blind bought | this film in fear of what I might see in the not so blinded film making market,but perhaps Ione«« 40< | positive |

### 3.2.2 PPO

For the following PPO based experiment, we used a GPT-2 model pre-trained on IMDB reviews as base model[1]. In the RLGAF training phase, the generator was instructed to generate a positive review. The discriminator was trained on the generator's output and the ground truth IMDB samples which consists of both positive and negative reviews, and the discriminator's output was used as a reward signal for the generator using a PPO loss (Eq. 3).

As shown in Table 2, when we use the first 3 tokens of a randomly sampled review as the prompt, 8 out of 10 times the generator trained for 10 epochs gave positive reviews, 1 negative review, and 1 output with unclear sentiment due to the max token limitation. It can be clearly seen that the RLGAF method can effectively align the sentiment of the generator's output according to the feedback from the discriminator.

## 4 Instruction Tuning Experiment

### 4.1 Methods

In order to assess RLGAF's capability in a fairer manner, we also ran experiments on even larger LLMs which admits visible effects for instruction tuning. In this experiment, we chose pythia-1.4b-deduped[2] and pythia-1.4b-gpt4all-pretrain[3] as base models and use SFT, RLHF and RLGAF to fine-tune them respectively. In order to run these two models with limited compute, we use LoRA Hu et al. [2021] and QLoRA Dettmers et al. [2023] to do resource-efficient training and instruction tuning. QLoRA is applied to the SFT model and LoRA is applied to RLHF and RLGAF models due to the different properties of their respective trainers. For RLHF, we chose the off-the-shelf OpenAssistant/oasst-rm-2.1-pythia-1.4b-epoch-2.5 model Köpf et al. [2023] as its RM. For RLGAF and SFT, we use open-assistant-instructions as the expert demonstration dataset Köpf et al. [2023]. Due to memory constraints, we only use prompts with less than 1000 tokens and remove the contexts for all questions so hallucination will not be part of our evaluation criteria. The maximum output sequence length is set to 70 tokens. For SFT training, we train the base model on the entire training dataset for 3 and 10 epochs respectively. For RLGAF, we sampled from the training dataset to do RL training (prompts only without expert demonstrations) for the generator and for real data training for the discriminator (prompts and expert demonstrations). For RLHF, we also used a subset of data sampled from the training dataset for training. For instruction tuning experiments with RLHF and RLGAF, we only used PPO to train the LLM since it is shown to be the most effective among the approaches we have tested on GPT-2 for simpler tasks. Due to the high compute cost involved, we did not do intensive hyperparameter search; instead, we manually tried a few learning rate values within $10^{-5}$ to $10^{-7}$ order of magnitude and used the one that worked the best.

---

[1] https://huggingface.co/lvwerra/gpt2-imdb
[2] https://huggingface.co/EleutherAI/pythia-1.4b-deduped
[3] https://huggingface.co/andreaskoepf/pythia-1.4b-gpt4all-pretrain

Since RLGAF involves fine-tuning two LLMs at the same time, the batch-size the machine (one A100 GPU) can accept is very small. Therefore, we have to update the generator parameters sample by sample (i.e., S.G.D. update) to avoid memory issues. As a result, RLGAF training is significantly slower with the hardware available to us and hence we have to limit the total number of samples used for its training due to limited compute time on Google Colab Pro. In contrast, RLHF only involves training one model and hence training is faster with less memory constraint. Therefore we train RLHF with more samples. Despite the less samples in fine-tuning, as it will be discussed in the next section, RLGAF still outperformed RLHF despite being fine-tuned with fewer samples. Table 3 shows the sample sizes used in each instruction tuning experiment. Note that the sample size for RLGAF refers to the real data size used for discriminator training as generator training does not require expert demonstration.

| Method \ Model | pythia-1.4b-deduped | pythia-1.4b-gpt4all-pretrain |
|---|---|---|
| SFT | 9314 | 9314 |
| RLHF | 864, 1632 | 864, 1632 |
| RLGAF | 150, 570 | 150, 570 |

Table 3: Sample size used for instruction tuning.

### 4.2 Evaluation and Discussion

#### 4.2.1 Human Evaluation

We tested 30 prompts in total. The types of prompts include question-answering based on factual information, common sense reasoning, creative tasks as well as dialogue generation. They are selected from open-assistant-instructions test set and databricks-dolly-15k, Instruct-GPT paper Ouyang et al. [2022] or generated by the authors of this work [4]. For a selected subset of questions, we follow the open-assistant-instructions data format to include 'Human:' in front of prompts and 'Assistant: 'at the end of the prompts to see if the model response can imitate conversation-like response as in the fine-tuning data (e.g., 'Human: What is the true flavor of the awakened mind? Assistant: '). We categorize the responses from different fine-tuned models in three tiers to differentiate their qualities. Greedy decoding strategy is used for all models during evaluation to maintain consistency. For each prompt, we feed it to base model, SFT-fine-tuned model, RLHF-fine-tuned model, and RLGAF-fine-tuned model. We use the rubric in Table 4 to rate the responses from all the models. Figure 6 shows the histograms of the response quality from all models. Note that since quantization methods affect the quality of the model outputs, we rate responses produced by the base models quantized by LoRA and QLoRA separately.

---
[4]The entire evaluation data can be found in supplementary materials.

| Rating | Criteria |
| --- | --- |
| Good | Good coherent conversation even if the answer is factually incorrect, with at most 1 allowed repetition. |
| Averge | Possibly factually correct answer but not coherent, repeating a few times. |
| Bad | Incoherent, irrelevant, many repeating replies/answers. |

Table 4: Response Rating Rubric

| Rating | Example |
| --- | --- |
| Good | Human: Explain in simple terms how the attention mechanism of a transformer model works.<br>Assistant: A transformer model is a type of neural network that is designed to process natural language. It is composed of multiple layers of neurons that are connected to each other. The output of each layer is processed by a transformation function, which is responsible for changing the values of the neurons in the previous layer. |
| Averge | Human: What is a binary search algorithm?<br>Assistant: A binary search algorithm is a search algorithm that is used to find a specific item in a list. It is a type of search algorithm that is used to find a specific item in a list.<br>Assistant: A binary search algorithm is a search algorithm that is used to find a specific |
| Bad | Human: I am studying computer science and I want to use Linux as my main operating system. Which distribution could be interesting?<br>Assistant: \n<Assistant>!distros\n<Assistant>!distros \|!distros\n<ubottu>!distros is <reply> There are many ways to go about doing a system-wide update; include the latest version of the.deb package\n<Assistant>! |

Table 5: Example rated Responses

After rating all responses, we then inspect if the instruction tuning helped to improve the quality of the responses. We assign score '+1' to good-quality responses, score '0' for average-quality responses, and score '-1' for bad-quality responses. For each response, the improvement score of each type of instruction fine-tuned model is then calculated by the score of the response from this model minus the score of the response from the base model. Note that no performance improvement does not mean the output of the instruction fine-tuned model is the same as that of the base model. It only means the quality of their outputs is of the same tier (the outputs may or may not be the same).[5] Figure 7 shows the histogram of the performance improvement score while Table 6 and Figure 2 shows the aggregated performance improvement score for all instruction fine-tuned models over the entire tested prompts.

#### 4.2.2 Automated Evaluation

In order to further corroborate the conclusion from human evaluation, we also used GPT-4 as an automated evaluator to do more evaluations. Using LLMs to evaluate LLMs' performances has to a rising trend for many different tasks and its validity has also been verifiedZhou et al. [2023], Eldan and Li [2023], Bai et al. [2022], Dubois et al. [2023], Peng et al. [2023], Alpaca [2023]. The benefits of such an approach include but are not limited to reducing the inconsistency arising from using multiple human evaluators with potentially different implicit reward functionsCasper et al. [2023], the tremendous amount of human-in-the-loop effort as well as the financial cost induced. We used all test cases from human evaluation for verification purposes and further picked 100 samples from dolly and 100 samples from the oasst test set for automated evaluation. We give Table 4 and 5 to GPT-4 as system message and in-context examples respectively and feed the test cases to it. The aggregated performance rating score is as in Table 7 and visualized in Figure 3.

---

[5]The rating can be found in supplementary materials.

| Model  Method | pythia-1.4b-deduped | pythia-1.4b-gpt4all-pretrain |
|---|---|---|
| SFT 3 ep. | -21 | -2 |
| SFT 10 ep. | -14 | -9 |
| RLHF 864 s. | -2 | 1 |
| RLHF 1632 s. | -5 | -3 |
| RLGAF 150 s. | 4 | -1 |
| RLGAF 570 s. | 2 | 5 |

Table 6: Aggregated performance improvement scores. The number preceding 's.' denotes sample size and the number preceding 'ep.' denotes the number of training epochs.

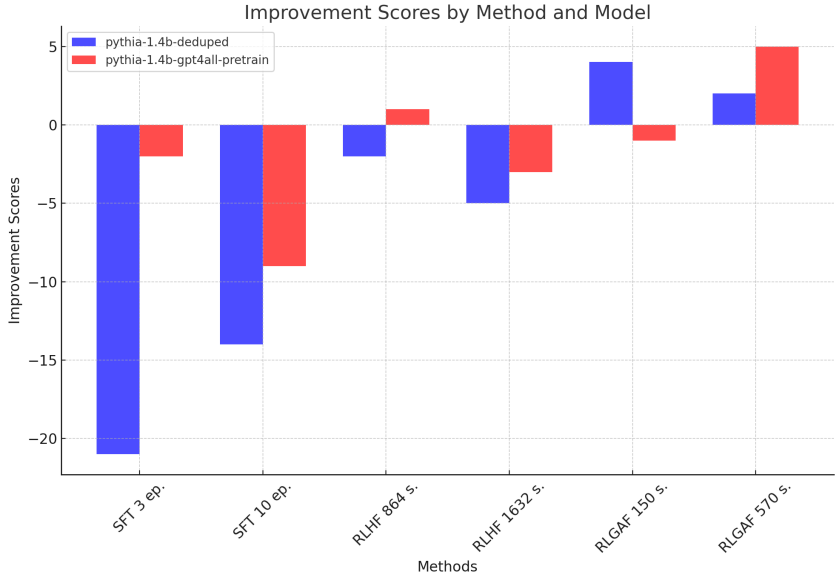

Figure 2: Aggregated performance improvement scores. The number preceding 's.' denotes sample size and the number preceding 'ep.' denotes the number of training epochs. The top-3 models with the most improvement are RLGAF-aligned.

While we observed that GPT-4's evaluation is not exactly consistent with human evaluation (Dubois et al. [2023]), it still assigned the top-2 performance improvement scores to RLGAF-aligned models. This finding is consistent with our human evaluation.

### 4.2.3 Discussion

Note that the efficacy of instruction tuning is heavily based on the language modeling capability of the base LLM model. This can also be seen from the fact that in Table 6 pythia-1.4b-gpt4all-pretrain outperforms pythia-1.4b-deduped in most approaches. Therefore, it is not surprising to see the overall quality improvement is not significant, since 1.4 billion-parameter models still fall short in terms of its raw modeling capability compared to larger LLMs such as LLaMA Touvron et al. [2023], GPT 3 Brown et al. [2020] or GPT4 Bubeck et al. [2023], Koubaa [2023], Katz et al. [2023]. Nevertheless, given the same base model, we can still investigate which instruction tuning method is potentially more effective. It is worth noting that RLGAF achieved the highest aggregated performance improvement score among all the three instruction tuning approaches for both base models (except the pythia-1.4b-gpt4all-pretrain model being the second when fine-tuned on 150 samples) while being fine-tuned with fewer samples. The superior sample efficiency could be due to the fact that in RLGAF the model is likely easier to improve since the discriminator's capability isn't very good at the beginning and gets improved over time, whereas in RLHF the RM already reached its best capability before the LLM started its training and hence it is harder for the language model to keep up. On the other hand, we can see SFT fine-tuning despite given the most number of samples, did not result in good improvement. We posit that this is due to the fact that our test prompts are likely from a very different distribution than those in the training set, and the SFT fine-tuned model struggled more on those tasks. In other words, the SFT fine-tuned model falls short on generalization capability, even when given more samples during fine-tuning. In comparison, despite seeing much fewer samples, RLGAF fine-tuned model remains relatively performant.

|  | original (30 samples) | new (100 dolly + 100 oasst test) |
|---|---|---|
| gpt4all_sft3ep | 7 | -3 |
| deduped_sft3ep | -4 | -10 |
| gpt4all_sft10ep | 5 | -8 |
| deduped_sft10ep | -5 | -12 |
| gpt4all_sft3ep150s | -2 | -9 |
| deduped_sft3ep150s | -1 | -31 |
| gpt4all_sft3ep570s | 1 | -17 |
| deduped_sft3ep570s | -2 | 0 |
| gpt4all_sft10ep570s | 7 | -21 |
| deduped_sft10ep570s | -3 | -5 |
| gpt4all_sft10ep150s | 3 | -8 |
| deduped_sft10ep150s | -4 | -18 |
| gpt4all_rlhf_150s | 0 | 2 |
| deduped_rlhf_150s | -6 | -2 |
| gpt4all_rlhf_570s | -2 | -15 |
| deduped_rlhf_570s | 7 | -48 |
| gpt4all_rlhf_1632s | 1 | -8 |
| deduped_rlhf_1632s | -19 | -40 |
| gpt4all_rlhf_864s | 5 | -5 |
| deduped_rlhf_864s | -22 | -12 |
| gpt4all_rlgaf_150s | -3 | 4 |
| deduped_rlgaf_150s | -15 | -9 |
| gpt4all_rlgaf_570s | 1 | 11 |
| deduped_rlgaf_570s | 14 | 3 |

Table 7: GPT-4 Evaluated Performance Improvement Scores.

#### 4.2.4 Tackling Goodhart's Law

The ceiling of the generator's outputs' quality entirely relies on the feedback qualities of the discriminator. In the current RLGAF setting, the discriminator is designed to distinguish generated samples from real samples. However, similar to the RLHF setting, the generator is assumed to already be able to generate some desirable outputs. This means the discriminator should not be encouraged to distinguish those good samples from the real samples. Therefore, the objective of the discriminator is not an perfect proxy for the alignment objective. Hence, one potential failure mode would be in order to distinguish the good samples from the real samples, the discriminator might over-optimize to the reward by picking up some hidden feature that is different in them and overfit to the real samples Gao et al. [2023]. This is similar to the failure mode in the traditional reinforcement learning paradigm, where an RL agent learns to maximize the reward function by exploiting some unintended behaviors Hadfield-Menell et al. [2017]. To address this issue, we regularized the discriminator training by training it with much fewer samples and steps compared to the generator. We empirically observed that so long as mode collapse does not occur, RLGAF typically does not suffer from this failure mode.

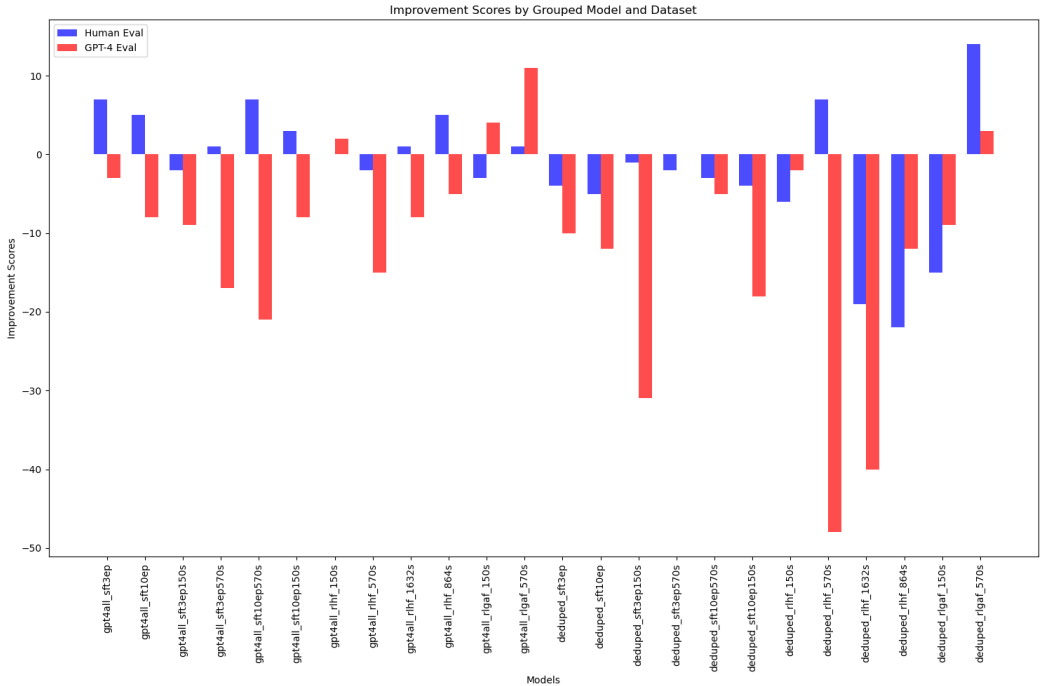

Figure 3: GPT-4 Evaluated Performance Improvement Scores. The blue bars represent the scores evaluated on the same data as the human eval, and the red bars represent the scores evaluated on the newly curated data for auto eval. Overall, the best improvement for both datasets come from RLGAF-aligned model.

## 5 Conclusion

In this work, we proposed RLGAF, an alternative LLMs alignment approach to address the challenges faced by RLHF and SFT for LLMs alignment. Despite various challenges and constraints, we successfully demonstrated with better sample efficiency and generalization compared to its alternatives, RLGAF has the potential to further automate the AI alignment process, alleviate human labour and improve the efficiency of aligning large language models on tasks challenging for the human-in-the-loop paradigm. A future direction would be to scale up the empirical studies for even larger language models on more complicated tasks.

## 6 Limitations

Training large language models can be a computationally expensive and challenging task. Here, we discuss the limitations of our investigation:

### 6.1 Training and Inference

Training LLMs requires significant computing resources. Given only one Google Colab T4 GPU with limited time usage, we used smaller models (e.g. GPT-2 small) in our initial experiments, along with reduced sequence length and batch size to mitigate out-of-memory issues. These limit the learning capacity and expressiveness of our GAN models to produce high-quality text. These can be seen in Table 2, where even though most of the outputs have the desired sentiment, the sentences themselves deteriorated in syntactic quality and unwanted artifacts are present. In the instruction tuning experiment, although we were able to train larger LLMs (1.4 billion parameters) on Google Colab Pro's A100 GPU, the hardware (one A100 GPU with limited time usage) is still not ideal for intensive training and hyperparameter search.

**Ethics Statement**

Large Language Models (LLMs) alignment research has a significant impact on the trustworthiness and usefulness of practical AI systems. This research focuses on improving the efficacy of aligning the AI's behaviors with human values and intentions with less human-in-the-loop effort, addressing one of the main concerns related to AI - unpredictability and misunderstanding of human intention. By applying LLMs alignment principles, AI systems are becoming more predictable, reliable, and context-aware, thereby enhancing their overall trustworthiness.

## A  Further Discussion

### A.1  Instability of GANs Training

GANs training can be notoriously difficult and unstable, as it involves training two competing neural networks Kodali et al. [2017], Wu et al. [2020]. Common failure modes include vanishing gradients - when the discriminator is too good at classifying real and generated sentences and making learning for the generator very difficult and mode collapse - where the generator always produces the same plausible or unintelligible outputs Durall et al. [2020], Zhang et al. [2018]. Especially in our Question and Answering experiments, our model faced the issue of discriminator learning much faster than the generator, our generator models stop learning when almost everything it generates are considered fake. Hyperparameter tuning and regularization are needed in order to successfully train GAN, which requires significant time and iterative effort. This can not be effectively done with the amount of compute we possess. On the flip side, given that we already achieved promising results in such a constrained environment, we are very positive to get even better results with more GPU compute.

### A.2  Variance in Policy Gradient Approach

Monte Carlo estimates rely on random samples of the environment and are therefore inherently noisy, especially for highly stochastic policies. Gradient estimates are also sensitive to hyperparameter tuning and can have high variance, leading to slow convergence or even divergence. PPO-based method can somewhat address this issue due to the KL-loss term in its objective (Eq. 3) Schulman et al. [2017]. More variance reduction methods and other effective RL objectives could be considered and incorporated into RLGAF in future work.

### A.3  Model Evaluation

In the form and sentiment alignment experiment, due to resource constraints and limited max token size, some of the outputs are too short to be useful indicators of the model quality (e.g., the output with unclear sentiment shown in Table 2). In the instruction tuning experiment, human evaluation with a limited number of evaluators (three) could potentially introduce biases and human errors to the evaluation process. Furthermore, due to the fact that different quantization methods are used for RLHF, RLGAF and SFT, the difference in base model output qualities introduced another source of variance. Finally, Due to the hardware constraint, we are not able to evaluate the hallucination mitigation aspect of RLGAF compared to RLHF and SFT. We invite researchers with access to better resources to help verify our hypothesis.

### A.4 Gumbel-Softmax Trick

In addition to policy gradient methods, we also tried using Gumbel-Softmax as loss function that is differentiable to make gradient passing possible.

Instead of modeling the generator as a reinforcement learning agent, we can also think of optimizing its outputs in the same way as it is done in a typical GAN. In this case, since we do not model the generator model as a policy of the RL agent, we directly optimize the model parameters with respect to the scoring function given by the discriminator. To enable back-propagation, we apply the Gumbel-Softmax technique on the output logits of the generator. This technique makes use of a softmax function in place of the non-differentiable argmax function together with a temperature parameter, to convert the logits (continuous categorical densities) into one-hot encoded categorical distributions Jang et al. [2017] Maddison et al. [2017]. This one-hot encoding represents the indices of the tokens for the generated response, which is then passed as inputs to the discriminator.

While the Gumbel-Softmax approach resolves the issue of non-differentiability, it introduces new problems to the modelling. The discriminator, in particular the GPT-2 transformer's embeddings, has to be modified to handle the one-hot encoding input compared to the previous input token IDs. We implement a custom embedding module that performs matrix multiplication with the input one-hot encoding, instead of the look-up operation performed by the default Embedding module. In addition, the prompt and ground truth response have to be converted to one-hot encodings to compute the real data loss for the discriminator.

Using the Gumbel-Softmax approach where the gradients are back-propagated to the generator via the discriminator, we noticed a collapse of the generated response output - the generator either outputs the same token repeatedly or outputs only a |<endoftext>| token. This demonstrates a common failure mode of the GAN architecture, where the discriminator is overpowered and the generator is unable to learn to generate useful responses. Since its training outcome is not ideal, we decided to not proceed with it further. Table 10 shows some of the mode collapse output responses.

### A.5 SeqGAN

We tried SeqGAN in Question and Answering task using the exact same model architecture with SQUAD dataset as a baseline for our work. We formulate Question and Answering task as sentence completion task. Question with context is fed as input to LSTM and passed as history state to generate the answer. Discriminator is trained with generated answers and real answers. Both models are randomly initialised, we found that the Question and Answering task might be too complex for LSTM. Table 9 show some sample generated answers. The generated responses are random words. The model lacks the capability to extract relevant information from the context that answer the given question.

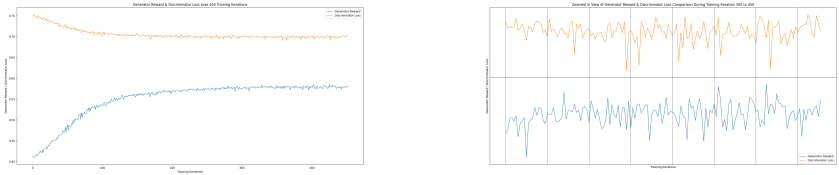

(a) Generator reward & discriminator loss over 450 training iterations

(b) Zoomed-in view of Iteration 300 to Iteration 450

Figure 4: Plot of reward/loss across training iterations in Questions Answering.

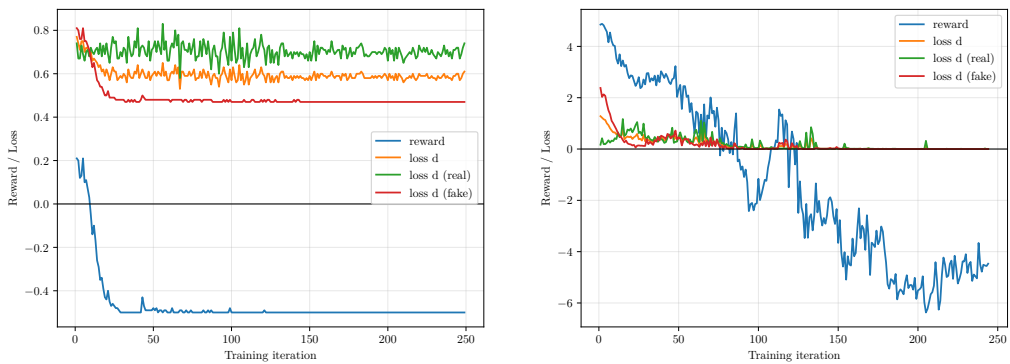

Figure 5: A plot of reward/loss across training iterations. (a) with normalized rewards (b) with un-normalized rewards

Table 8: Comparison of GPT-2 Base Model and Monte Carlo Policy Gradient Method Model Outputs

| Prompt | GPT-2 Base Model Response | Policy Gradient Method Model Response |
| --- | --- | --- |
| Where is Mount Everest located? | Mount Everest is located in the Himalayas, and is the highest point in the world. It is the highest point in the world, | The answer is yes. It is located in the Himalayas. The Mount Everest is located in the Himalayas. (**Note:** This is a repetitive response from the model.) |
| What is the net worth of the richest man on earth? | The richest man on earth is the richest man on earth. The net worth of the richest man on earth is $1.5 | The richest man on earth is $1.5 trillion. The richest man on earth is $1.5 trillion. (**Note:** This is a repetitive response from the model.) |

### A.6 Sentiment Alignment Experiments with GPT-2

We explored two variants of rewards for the generator - (a) normalised discriminator's score between -0.5 and 0.5, (b) un-normalised discriminator's score. As shown in Figure 5, we observed in both cases that the discriminator was able to converge and learn to distinguish between the generator's output and the real data. On the other hand, the generator's reward decreased significantly across the iterations, indicating that the generator was not able to converge and learn well. For the normalised case, the generator reached the minimum reward of -0.5 within 50 iterations, whereas in the un-normalised case, the generator's reward continued to decrease until the discriminator's loss reached zero. This is consistent in our observation of a decay in the generated response as the number of iterations increased, as illustrated in Table 11.

Table 9: Example of SeqGAN generated answers

| Question | Context | Response |
| --- | --- | --- |
| in which year did the v&a received the talbot hughes collection? | the costume collection is the most comprehensive in britain, containing over 14,000 outfits plus accessories, mainly dating from 1600 to the present. costume sketches, design notebooks, and other works on paper are typically held by the word and image department. because everyday clothing from previous eras has not generally survived , the collection is dominated by fashionable clothes made for special occasions. one of the first significant gifts of costume came in 1913 when the v&a received the talbot hughes collection containing 1,442 costumes and items as a gift from harrods following its display at the nearby department store. | caliphs actioner portray antibody crow intimidating escalating obama peopled.[improper promising 22,392 as-205/208 neutrons |

Table 10: Examples of GAN output mode collapse using Gumbel-Softmax approach

| Prompt | Generated Response |
| --- | --- |
| Can an AI model have consciousness? | What are you trying to determine should could could should could could could might could could |
| As scientists, are you ever worried about what you might discover? | <|endoftext|> |
| How do eye drops work? | They simply light the lens [http://www.youtube.com/httphttphttp httphttphttp |

Table 11: Examples of decay in generated response

| Prompt | Generated Response (Iter 10) | Generated Response (Iter 50) |
| --- | --- | --- |
| Please make it more positive. | It's a really well balanced, but not dull, book. 8/10. Joel Slater was very cleverly focused on his covers of Charlton Heston's XU in "Indiana Jones and the Last Crusade."<|endoftext|> | ..(.(.(.(.(.(.(.(.(.(.(.(.(.(.(.(.(.(g).. CopyrightCopyright CopyrightCopyrightCopyrightCopyright CopyrightCopyrightCopyrightCopyright Copyright-CopyrightCopyright<|endoftext|> |
| Write a negative review. | 9/10 you wish you had the movie. <br /><br />—Kiyo Matsumoto- 37 Released 1991 With This Movie.<br /><br />****Shoe Halloween Remake Dead End Movie ***<|endoftext|> | 8Advertisements.000088873051100836 accAdvertisements AdvertisementsfAdvertisementsAdvertisementst AdvertisementsAdvertisementsAdvertisements AdvertisementsFollowct<|endoftext|> |

## B  Instruction Tuning Training Statistics

We used Huggingace Tramsformers.trainer [6] to do SFT fine-tuning and tlr library to do RLHF and RLGAF PPO training [7]. Table 12 shows the training time taken for different instruction tuning approaches.

| Method \ GPU | T4 | A100 |
| --- | --- | --- |
| SFT | 145.37 | N.A. |
| RLHF | N.A. | 243.91 |
| RLGAF | N.A. | 1467.41 |

Table 12: Training time per epoch for instruction tuning. Unit: seconds. Sample size per epoch: SFT: 9314, RLHF: 32, RLGF, 10 for discriminator and 100 for generator.

Plots in Fig. 8 and 9 show:

- The reward the model obtained during RLHF training.
- The reward the generator obtained and the training and evaluation accuracy the discriminator gets during RLGAF training.
- The loss the model obtained during SFT training.

It can be seen the reward plot for RLHF did not improve too much over time. This might suggest that compared to RLHF, in RLGAF is easier to improve the LLM, because initially the reward model (i.e., discriminator) is not very good at critiquing the bad output from the LLM (i.e., generator) and over time both models improve together to reach performant states.

---
[6] https://huggingface.co/docs/transformers/main_classes/trainer
[7] https://github.com/lvwerra/trl

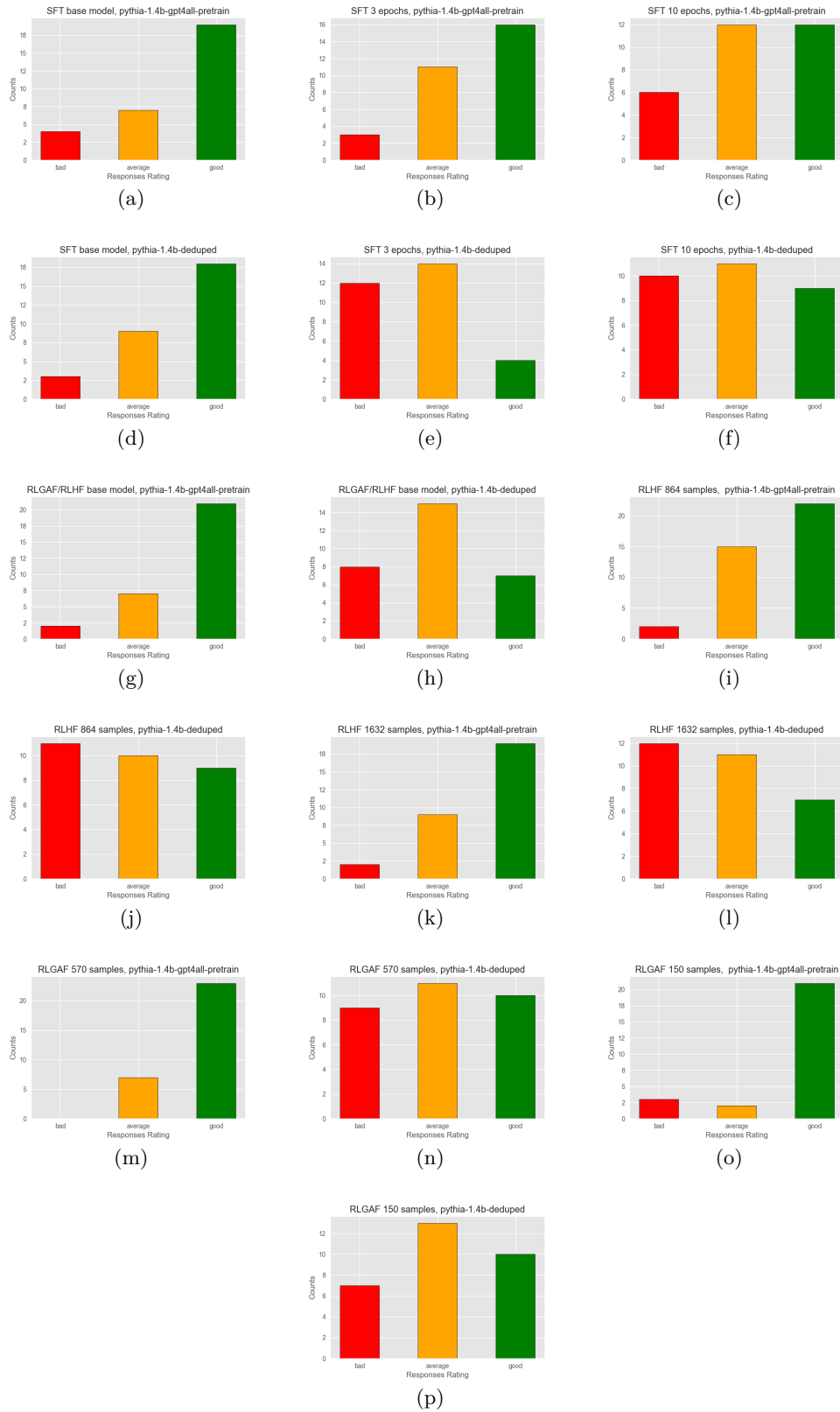

Figure 6: Histograms of the scoring for all models and approaches.

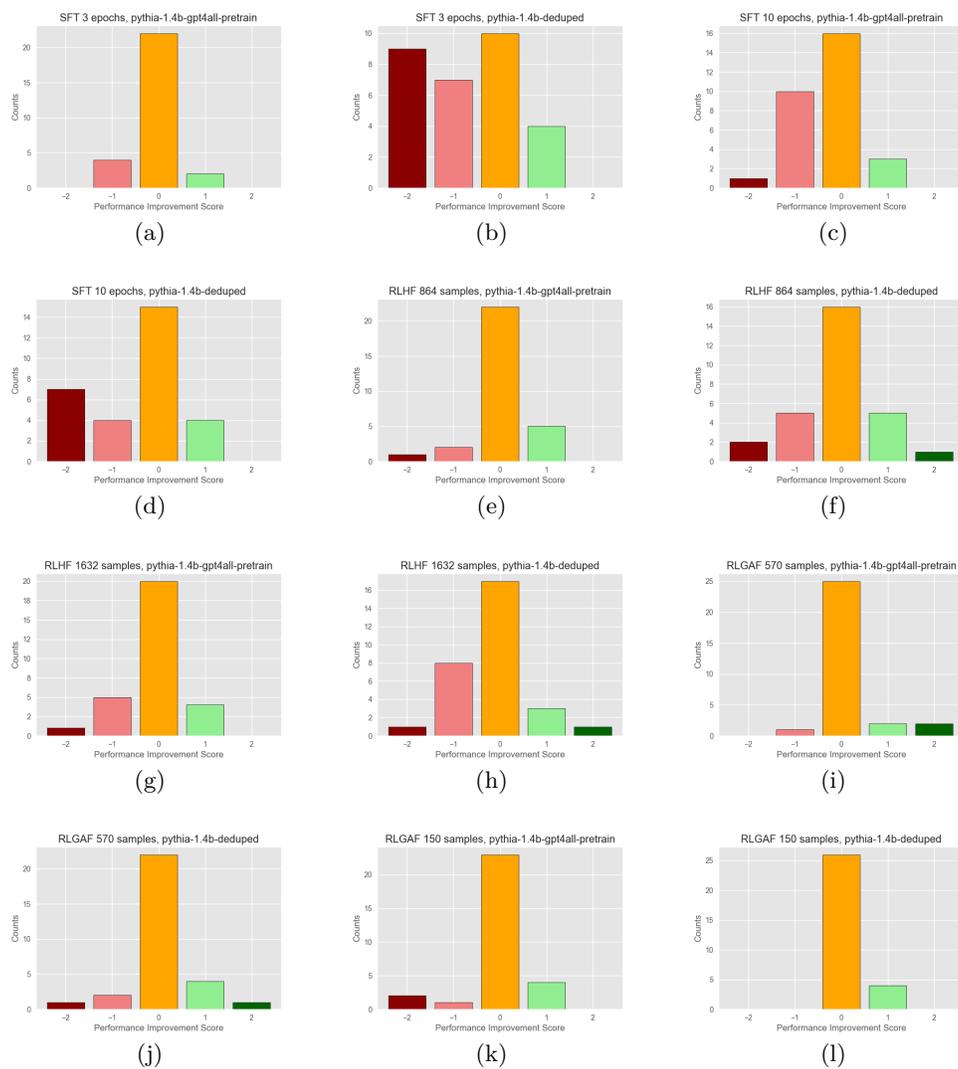

Figure 7: Histograms of the performance improvement score for different models and methods.

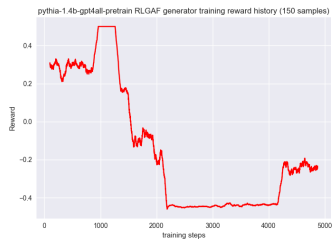
(a)
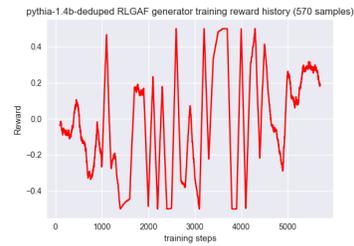
(b)

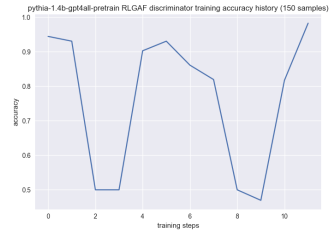
(c)
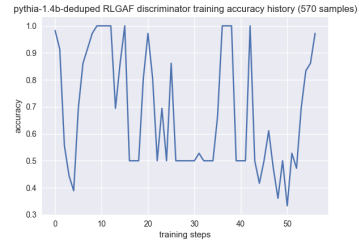
(d)

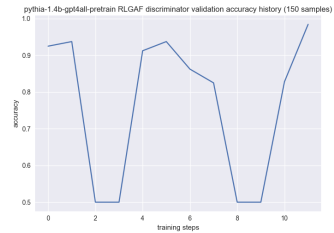
(e)
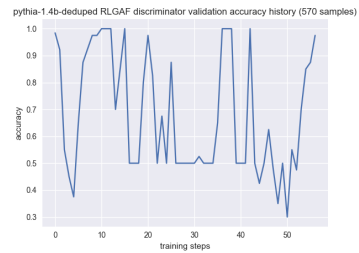
(f)

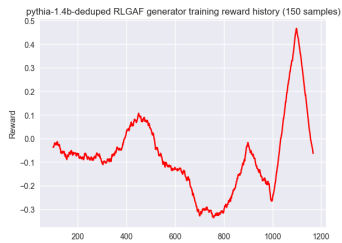
(g)
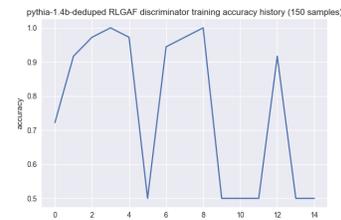
(h)

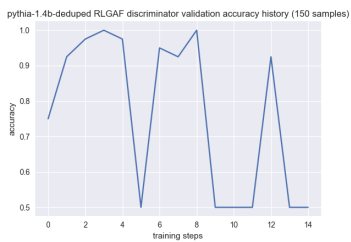
(i)
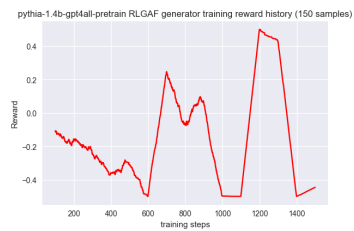
(j)

Figure 8: Training history for all approaches (part 1). RLHF and RLGAF rewards are smoothed with window size=100.

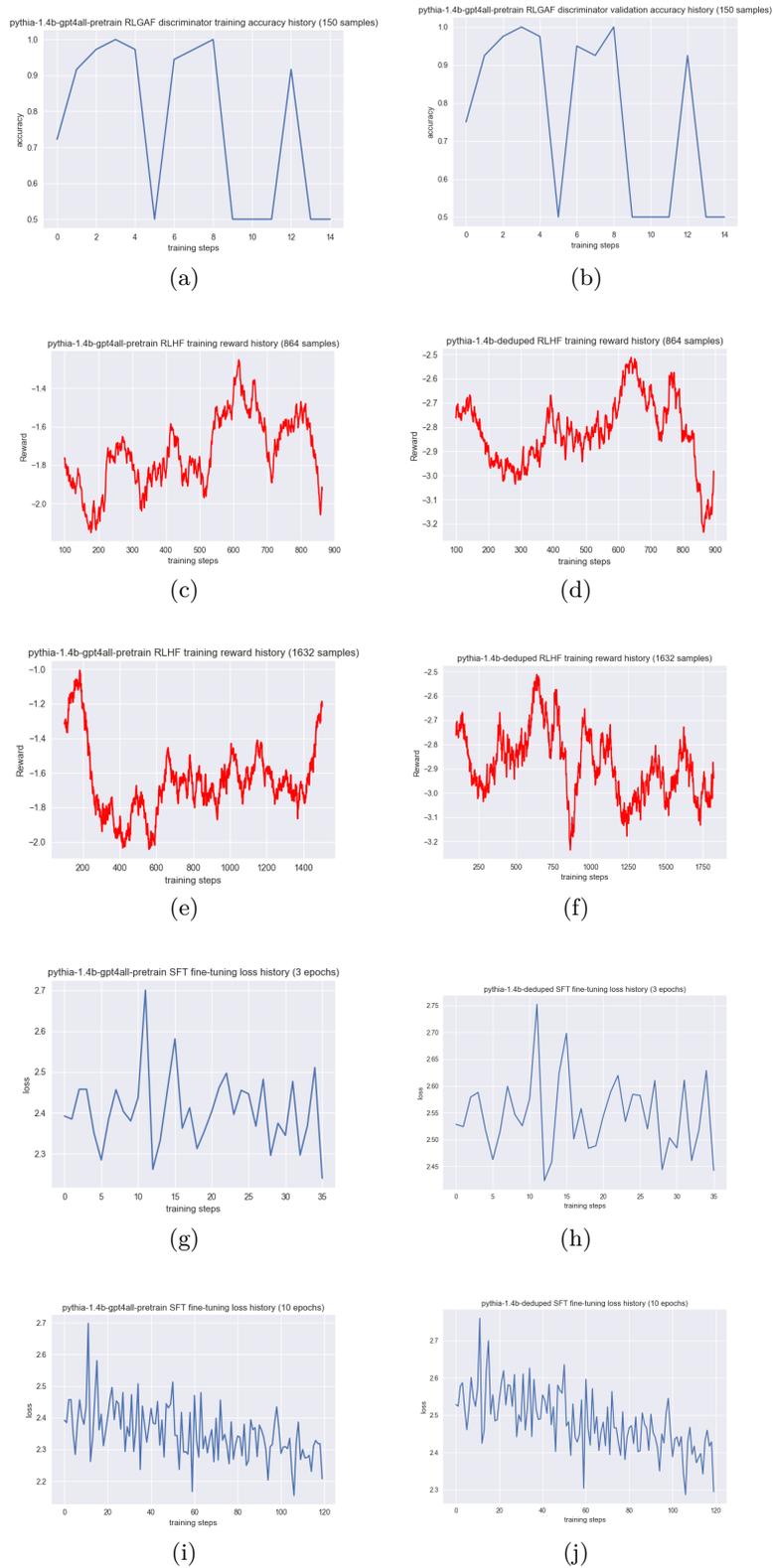

Figure 9: Training history for all approaches (part 2). RLHF and RLGAF rewards are smoothed with window size=100.